\documentclass[sigconf,nonacm]{aamas}

\usepackage{balance} % for balancing columns on the final page
\usepackage{algorithm}
\usepackage[noend]{algpseudocode}
\usepackage{subcaption}
\usepackage{mathtools}
\usepackage{multirow}

\setcopyright{ifaamas}
\acmConference[AAMAS '25]{Proc.\@ of the 24th International Conference
on Autonomous Agents and Multiagent Systems (AAMAS 2025)}{May 19 -- 23, 2025}
{Detroit, Michigan, USA}{A.~El~Fallah~Seghrouchni, Y.~Vorobeychik, S.~Das, A.~Nowe (eds.)}
\copyrightyear{2025}
\acmYear{2025}
\acmDOI{}
\acmPrice{}
\acmISBN{}

\acmSubmissionID{1383}

\title[AAMAS-2025 Formatting Instructions]{Personalized Federated Learning with Adaptive Feature Aggregation and Knowledge Transfer}

\author{Keting Yin}
\affiliation{
  \institution{Zhejiang University}
  \city{Hangzhou}
  \country{China}}
\email{yinkt@zju.edu.cn}

\author{Jiayi Mao}
\affiliation{
  \institution{Zhejiang University}
  \city{Hangzhou}
  \country{China}}
\email{jiayimao2000@zju.edu.cn}

\begin{abstract}
Federated Learning(FL) is popular as a privacy-preserving machine learning paradigm for generating a single model on decentralized data. However, statistical heterogeneity poses a significant challenge for FL. As a subfield of FL, personalized FL (pFL) has attracted attention for its ability to achieve personalized models that perform well on non-independent and identically distributed (Non-IID) data. However, existing pFL methods are limited in terms of leveraging the global model's knowledge to enhance generalization while achieving personalization on local data. To address this, we proposed a new method \textit{personalized \textbf{Fed}erated learning with \textbf{A}daptive \textbf{F}eature Aggregation and \textbf{K}nowledge Transfer} (\textbf{FedAFK}), to train better feature extractors while balancing generalization and personalization for participating clients, which improves the performance of personalized models on Non-IID data. We conduct extensive experiments on three datasets in two widely-used heterogeneous settings and show the superior performance of our proposed method over thirteen state-of-the-art baselines.
\end{abstract}


\newcommand{\BibTeX}{\rm B\kern-.05em{\sc i\kern-.025em b}\kern-.08em\TeX}

\begin{document}

%%% The following commands remove the headers in your paper. For final 
%%% papers, these will be inserted during the pagination process.

\pagestyle{fancy}
\fancyhead{}

%%% The next command prints the information defined in the preamble.

\maketitle 

%%%%%%%%%%%%%%%%%%%%%%%%%%%%%%%%%%%%%%%%%%%%%%%%%%%%%%%%%%%%%%%%%%%%%%%%
\section{Introduction}

In recent years, many learning tasks, such as financial fraud detection~\cite{chen2021deep,alarfaj2022credit}, benefit from deep neural networks. To achieve satisfactory performance, model training requires vast amounts of data~\cite{lecun2015deep, krizhevsky2012imagenet}. However, in many scenarios, it is costly for a single entity to obtain enough data, especially when data is considered a valuable asset. Moreover, due to privacy concerns, the collection and central storage of data are often prohibited~\cite{regulation2016regulation,de2018guide}. Federated learning (FL), as a paradigm of collaborative learning, enables model training among distributed clients without accessing the raw data~\cite{fedavg,yang2019federated,li2020federated,zhang2021survey}. A typical federated learning process involves three main steps. First, the server selects a subset of clients in each round and distributes the global model to them. The selected clients then initialize their local models with the global model, update them using their local datasets, and send the updated models back to the server. Finally, the server aggregates the received local models to obtain a new global model, repeating this process until convergence. The aforementioned federated algorithm demonstrates excellent performance when data is independent and identically distributed (IID). However, in practice, data distributed across various clients is heterogeneous (non-independent and identically distributed, Non-IID) and unbalanced, making it challenging to learn a global model that meets the needs of all clients~\cite{fedavg,li2022federated,pfedme,zhu2021federated}.

To tackle with data heterogeneity, many researchers have focused on personalized federated learning (pFL). pFL aims to learn personalized models for each client rather than a single global model, treating data heterogeneity, which is typically seen as a limitation, as an advantage. Many methods have been applied to pFL: (1) Fine-tuning-based methods, \textit{e.g.}, FedAvg$+$Fine-tuning and Per-FedAvg~\cite{per-fedavg}, (2) Regularization-based methods, \textit{e.g.}, pFedMe~\cite{pfedme}, and Ditto~\cite{ditto}, (3) Clustering-based methods, \textit{e.g.}, ClusterFL~\cite{clusterfl}, (4) Model-splitting-based methods, \textit{e.g.}, FedPer~\cite{fedper}, FedRep~\cite{fedrep}, FedROD~\cite{fedrod}, and FedBABU~\cite{fedbabu}, (5) Personalized-aggregation-based methods, \textit{e.g.}, APFL~\cite{apfl}, FedFomo~\cite{fedfomo}, FedAMP~\cite{fedamp}, FedPHP~\cite{fedphp}, and FedALA~\cite{fedala}. In addition, many other algorithms for pFL have been proposed. Refer to ~\cite{tan2022towards, kulkarni2020survey} for more details.

Methods in (1), (2), and (3) typically initialize the local model with the global model, such as replacing the local model's parameters with those of the global model. These approaches introduce both relevant and irrelevant information into the local model, enhancing generalization but significantly reducing personalization. Methods in (4) decouples the model into a feature extractor (\textit{a.k.a}, the body) and a classifier head(\textit{a.k.a}, the header). The feature extractor is responsible for extracting low-dimensional feature representations (\textit{a.k.a}, feature embeddings) from the raw data, while the classifier head outputs classification results. In deep learning, learning a good feature extractor has been demonstrated to be crucial~\cite{bengio2013representation,fedrep}. However, current model-splitting-based methods do not fully leverage the knowledge gained during training to facilitate the transfer of useful knowledge, leading to limited performance improvements. Methods in (5) aims to obtain desired information from the global model or other clients' models during local aggregation~\cite{hanzely2020federated}. However, when the data distribution varies greatly, the adjustment of the personalization strategy can lead to fluctuations in the model performance. Additionally, this method may introduce extra system components, which incurs computation and communication overhead.

To address the aforementioned shortcomings, we propose the \textit{personalized \textbf{Fed}erated learning with \textbf{A}daptive \textbf{F}eature Aggregation and \textbf{K}nowledge Transfer} (\textbf{FedAFK}) to improve the performance of personalized models on Non-IID data. FedAFK consists of three simple but effective designs: model decoupling, knowledge transfer, and adaptive feature aggregation. During communication between the server and clients, only the feature extractor of the global model is transmitted, which reducing communication overhead through model decoupling. A term is incorporated into the training objective of the local feature extractor to transfer the knowledge contained in the global feature representation, enabling a better balance between the generalization of the classes across all data and personalization of the classes across local data. While training the global feature extractor and the local feature extractor, an aggregation coefficient $\mu$ is also learned. $\mu$ is used to explore the optimal combination of the two feature extractors, aiming to achieve better feature representation.

To evaluate the effectiveness of FedAFK, we conduct extensive experiments on three datasets in both pathological and practical settings. FedAFK outperforms thirteen state-of-the-art (SOTA) pFL methods. We summarize our contribution below:

\begin{itemize}
    \item We propose a novel pFL method FedAFK that adaptively aggregates the global feature extractor and the local feature extractor to enhance personalized model performance on Non-IID data by achieving better feature representation. 
    \item Within FedAFK, we transfer the global knowledge contained in representation to the local feature extractor, enabling a better balance between the generalization and the personalization. 
    \item We conduct extensive experiments on three datasets in two widely-used scenarios to show the effectiveness of FedAFK, which outperforms thirteen SOTA pFL methods in test accuracy without incurring additional communication overhead per round.
\end{itemize}

%%%%%%%%%%%%%%%%%%%%%%%%%%%%%%%%%%%%%%%%%%%%%%%%%%%%%%%%%%%%%%%%%%%%%%%%

\section{Related Work}

\subsection{Federated Learning with Data Heterogeneity}

As the standard algorithm in Federated Learning, FedAvg~\cite{fedavg} enables distributed clients to collaboratively learn a global model by aggregating their local models without accessing the raw data. However, FedAvg struggles to maintain robustness in heterogeneous FL settings~\cite{kairouz2021advances,zhao2018federated,zhu2021data}. Several methods are proposed to improve the global model learning. FedProx~\cite{fedprox} introduces a regularization term to the local training objective to maintain the stability of the learning process. SCAFFOLD~\cite{scaffold} corrects local update drift with control variates. Moon~\cite{moon} introduces a contrastive loss, which reduces the distance between the local model and the global model while increasing the distance between the current local model and the previous local model, thereby correcting the update direction of the local model. FedNova~\cite{fednova} redefines server-side weighting rules to eliminate target inconsistency and accelerate convergence.

\subsection{Personalized Federated Learning}

Personalized Federated Learning has emerged recently for tackling data heterogeneity~\cite{kairouz2021advances,wang2020optimizing,marfoq2021federated,cho2021personalized, achituve2021personalized,smith2017federated,mansour2020three}, which enables each client to have a customized model while benefiting from the federation. 

\textbf{Fine-tuning.} As the name suggests, FedAvg+Fine-tuning enables clients to fine-tuning their model locally  after completing the FedAvg process. Based on MAML~\cite{maml}, Per-FedAvg~\cite{per-fedavg} trains an initial model, and then each client performs several training steps on local data to obtain a personalized model. 

\textbf{Regularization.} pFedMe~\cite{pfedme} leverages Moreau envelopes to learn an additional personalized model for each client. Similar to pFedme, Ditto~\cite{ditto} trains additional personalized models with a proximal term to learn knowledge from the downloaded global model.

\textbf{Clustering.} ClusterFL~\cite{clusterfl} divides clients into different groups, where each group's clients have similar data distributions. This way, each group can perform more effective model aggregation within the group, thereby enhancing the performance of personalized models. 

\textbf{Model-splitting.} FedPer~\cite{fedper} and FedRep~\cite{fedrep} split the model backbone into a feature extractor and a classifier head. The feature extractor is shared through the server to learn generalizable knowledge, and the classifier head is used to fit local data and output classification results~\cite{kang2019decoupling,tian2020rethinking}. In FedROD~\cite{fedrod}, each client has a feature extractor and two classifier heads, enabling learning of both global and personalized tasks using separate heads. FedBABU~\cite{fedbabu} updates only the model's body during training, with the classifier head being randomly initialized and fine-tuned only during evaluation. 

\textbf{Personalized aggregation.} APFL~\cite{apfl} uses an adaptive weight to mix up the global and local model. This idea is similar to ours, but our method focuses on aggregating the parameters of the feature extractor rather than all model parameters, which enhances the personalization of layers closer to the output. FedFomo~\cite{fedfomo} enables each client to download other clients' models and calculates the client-specific weights for personalized aggregation. FedAMP~\cite{fedamp} uses an attention-inducing function to strengthen collaboration between similar clients. FedPHP~\cite{fedphp} aggregates the global model and the inherited private model linearly with a pre-defined hyperparameter. FedALA~\cite{fedala} conducts finer-grained model aggregation on the client side compared to APFL. 

\textbf{Other methods.} FedBN~\cite{fedbn} proposes a strategy for using batch normalization (BN) in the FL scenario to address the feature shift problem, while also leveraging BN to accelerate training process. FedCP~\cite{fedcp} focuses on the data, generating a conditional policy for each sample to separate the global information and personalized information in its features.

%%%%%%%%%%%%%%%%%%%%%%%%%%%%%%%%%%%%%%%%%%%%%%%%%%%%%%%%%%%%%%%%%%%%%%%%

\section{Problem Statement}

Suppose we have a central server and $n$ clients to collaboratively train personalized models without sharing raw data. For all clients, our training objective is:
\begin{equation}
    \min_{(w_1,\dots,w_n)\in\mathbf{W}}F(\mathbf{W}):=\frac{1}{n}\sum_{i=1}^nf_i(w_i)
\end{equation}
where $\mathbf{W}$ is the collection of all personalized models. $f_i(w_i)$ is the local objective for the $i$-th client:
\begin{equation}
    f_i(w_i):=\mathbb{E}_{(x_i,y_i)\in\mathcal{D}_i}\ell(w_i;x_i,y_i)
\end{equation}
where $\mathcal{D}_i$ is the data distribution and $\ell(w_i;x_i,y_i)$ denotes the loss, \textit{e.g.}, the cross-entropy loss, of the prediction on example $(x_i,y_i)$ made with model parameters $w_i$. 

%%%%%%%%%%%%%%%%%%%%%%%%%%%%%%%%%%%%%%%%%%%%%%%%%%%%%%%%%%%%%%%%%%%%%%%%

\section{Method}
%%%%%%%%%%%%%%%%%%%%%%%%%%%%%%%%%%%%%%%%%%%%%%%%%%%
\begin{figure*}[t]
	\centering
	\includegraphics[width=0.8\linewidth]{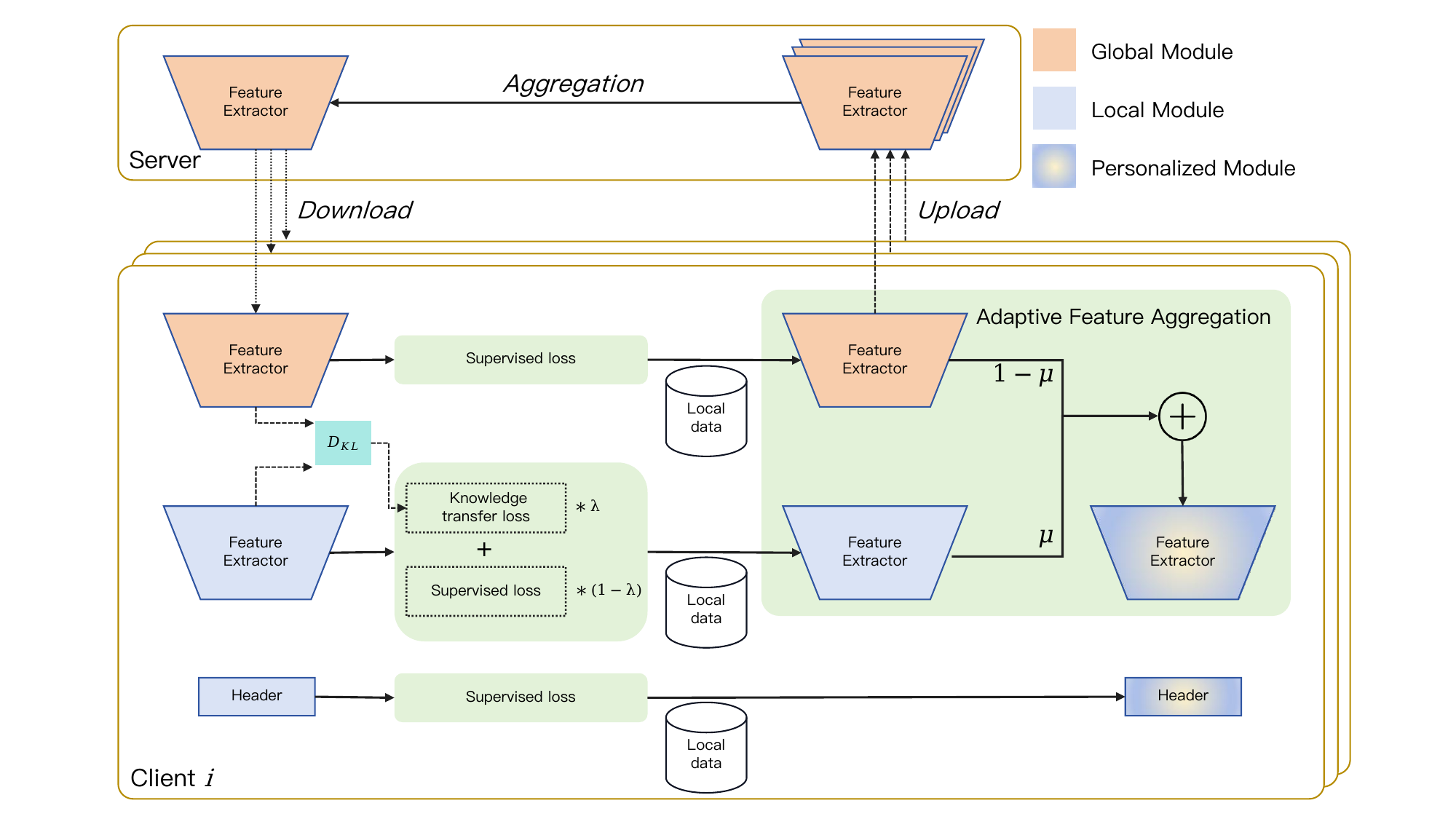}
	\caption{The workflow of FedAFK. $D_{KL}$ calculates the KL divergence of the feature representation obtained by the two feature extractors on the local data. $\lambda$ is a pre-defined weight of total loss. $\mu$ is a learnable adaptive coefficient.}
	\label{fig:illustrate}
        \Description{The workflow of our proposed FedAFK. }
\end{figure*}
%%%%%%%%%%%%%%%%%%%%%%%%%%%%%%%%%%%%%%%%%%%%%%%%%%%

\subsection{Adaptive Feature Aggregation}

Similar to FedPer~\cite{fedper} and FedRep~\cite{fedrep}, we decouple the deep neural network into the feature representation layers and the final decision layer~\cite{yosinski2014transferable,zhang2020decoupling,zhuang2021collaborative}. The feature representation layers, also known as the feature extractor $f:\mathbb{R}^D\rightarrow\mathbb{R}^K$, where $\mathbb{R}^D$ and $\mathbb{R}^K$ are the input space and low-dimensional feature space, respectively. In classification tasks, the final decision layer, referred to as the classifier head $g:\mathbb{R}^K\rightarrow\mathbb{R}^C$, where $\mathbb{R}^C$ is the label space. $D$, $K$ and $C$ are the dimension of their respective layers. Typically, $D$ is much larger than $K$, and $K$ is determined by the given model framework.

In FedAFK, we have a global feature extractor(parameterized by $\theta_G$), a local feature extractor(parameterized by $\theta_i$) and a local classifier head(parameterized by $\phi_i$), where the subscript $i$ denotes the client's id. To enhance the representation ability of the feature extractor, which serves as a common structure and has been shown to significantly improve personalized models~\cite{bengio2013representation,fedrep}, we introduce a learnable weight coefficient $\mu_i$. We use $\mu_i$ to compute the personalized feature extractor, aiming to achieve the optimal mixture of the local feature extractor and the global feature extractor:
\begin{equation}
    \theta_i = \mu_i\theta_i+(1-\mu_i)\theta_G
\end{equation}
For local training at the $t$-th communication round, we alternately learn the feature extractors and the classifier head. Specifically, after receiving the global feature extractor $\theta_G^{t-1}$ from the central server, we perform stochastic gradient decent to train models:
\begin{itemize}
    \item \textbf{Freeze $\phi_i$, Update $\theta_G$ and $\theta_i$} For the global model, we equip $\theta_G$ with a fixed random classifier head $\phi_{rd}$ and update it. For the personalized model, we freeze the classifier head and update $\theta_i$ with a combined loss $\ell_{total}$ that integrates knowledge transfer, which will be explained in Section~\ref{sec:kt}:
    \begin{equation}
        \theta_G^{t}\leftarrow\theta_G^{t-1}-\eta\nabla_{\theta_G}\ell_{sup}(\theta_G^{t-1},\phi_{rd};\mathcal{D}_i)
    \end{equation}
    \begin{equation}
        \theta_i^{t}\leftarrow\theta_i^{t-1}-\eta\nabla_{\theta_i}\ell_{total}(\theta_i^{t-1},\phi_i^{t-1},\theta_G^{t-1};\mathcal{D}_i)
    \end{equation}
    \begin{equation}
        \mu_i^{t}\leftarrow\mu_i^{t-1}-\eta\nabla_{\mu_i}\ell_{total}(\theta_i^{t-1},\phi_i^{t-1},\theta_G^{t-1};\mathcal{D}_i)
    \end{equation}
    where $\eta$ denotes the learning rate for updating, and $\ell_{sup}$ is the supervised loss, \textit{e.g.}, the cross-entropy loss.
    \item \textbf{Freeze New $\theta_i$, Update $\phi_i$} After getting the new personalized feature extractor, we train the classifier head for one epoch:
    \begin{equation}
        \phi_i^{t}\leftarrow\phi_i^{t-1}-\eta\nabla_{\phi_i}\ell_{sup}(\theta_i^{t-1},\phi_i^{t-1};\mathcal{D}_i)
    \end{equation}
\end{itemize}
With the adaptive weight $\mu_i$, we can find the find the optimal mixture of the local feature extractor and the global feature extractor, to obtain a better client-specific feature extractor, \textit{i.e.}, the personalized feature extractor. On one hand, the personalized feature extractor fuses the global feature extractor and the local feature extractor to achieve improved feature representation. On the other hand, this fusion also mitigates the risk of local over-fitting. When validating the performance of the personalized model, we perform inference using $w_i=\{\theta_i,\phi_i\}$.

\subsection{Knowledge Transfer}
\label{sec:kt}

Intuitively, a superior local feature extractor enhances effective adaptive feature aggregation. Therefore, we introduce an additional term in the loss function during the training of the local feature extractor to transfer the knowledge contained in the outputs of the global feature extractor, \textit{i.e.}, the global feature representation. This transfer helps achieve a better balance between generalization and personalization.
\begin{equation}
    \ell_{kt}(\theta_i,\theta_G)=D_{KL}(f_{\theta_i}(\mathcal{D}_i)||f_{\theta_G}(\mathcal{D}_i))
\end{equation}
where $D_{KL}$ refers to the KL-divergence, and $f(\mathcal{D}_i)$ denotes the feature representation on local data. The total loss $\ell_{total}$ is defined as:
\begin{equation}
    \ell_{total}=(1-\lambda)\ell_{sup}+\lambda\ell_{kt}
\end{equation}
where the coefficient $\lambda$ is the hyperparameter for knowledge transfer. 

It is worth mentioning that, in terms of learning global knowledge, incorporating knowledge transfer during the training of the local feature extractor seems to overlap with adaptive feature aggregation. To emphasize learning personalized knowledge from local data while maintaining the generalization, we select a reasonable $\lambda$ to mitigate this "redundancy" and strike a balance. 

\subsection{Workflow and Algorithm}
As illustrated in Figure~\ref{fig:illustrate}, the workflow of FedAFK includes the following steps:
\begin{enumerate}
    \item The server first broadcasts the aggregated global feature extractor to selected clients. 
    \item Client $i$ receives the global feature extractor and equips it with a fixed classifier head. Then, client $i$ trains the global feature extractor with supervised loss on the local dataset.
    \item Client $i$ calculates $D_{KL}$ as the knowledge transfer loss, defined as the KL divergence between the feature representations obtained from the global and local feature extractors. Then, client $i$ trains the local feature extractor with knowledge transfer loss and supervised loss on the local dataset.
    \item Client $i$ trains the weight coefficient $\mu_i$ and aggregates the global and local feature extractors accordingly. 
    \item Client $i$ trains the local classifier head with supervised loss on the local dataset.
    \item The updated local feature extractor and classifier head are combined to form a personalized model for inference, which also serves as the local model for the next round of training.
    \item Client $i$ uploads the updated global feature extractor to the server.
    \item Following FedAvg, the server aggregates the received global feature extractors to produce the global feature extractor for the next round. 
\end{enumerate}
The above steps are repeated until convergence. The detailed algorithm is presented in Algorithm~\ref{algo}.

%%%%%%%%%%%%%%%%%%%%%%%%%%%%%%%%%%%%%%%%%%%%%%%%%%%%%%%%%%%%%
\begin{algorithm}[ht]
        \caption{FedAFK}
        \label{algo}
	\begin{algorithmic}[1]
		\Require 
		$n$: number of clients; 
		$\rho$: client joining ratio;
		$T$: number of communication rounds; 
            $E$: number of local epochs;
		$\eta$: local learning rate;
		$[\mu_0,\dots,\mu_n]$: adaptive weights;
		$\lambda$: weight of the knowledge transfer loss.
		\Ensure 
		Personalized models $[w_1,\dots,w_n]$, where $w_i=\{\theta_i,\phi_i\}$.
        \State Server initializes the global feature extractor $\theta_G^0$.
        \State Each client initializes local feature extractor and classifier head, i.e., $\{\theta_i^0,\phi_i^0\}$ for the $i$-th client.
        \For{iteration $t=1,\dots,T$}
            \State Server samples a subset $S_t\leftarrow max(n\cdot\rho,1)$ clients.
            \State Server sends $\theta_G^{t-1}$ to clients in $S_t$. 
		    \For{Client $i \in S_t$ in parallel}
		        \State $\theta_{G,i}^{t-1}\leftarrow\theta_G^{t-1}$.
	            \For{local epoch $e=1,\dots,E$}
                        \State $\theta_{G,i}^{t}\leftarrow\theta_{G,i}^{t-1}-\eta\nabla_{\theta_{G,i}}\ell_{sup}$
                        \State $\theta_i^{t}\leftarrow\theta_i^{t-1}-\eta\nabla_{\theta_i}\ell_{total}$
                        \State $\mu_i^{t}\leftarrow\mu_i^{t-1}-\eta\nabla_{\mu_i}\ell_{total}$
                        \State $\theta_i^t = \mu_i\theta_i^t + (1-\mu_i)\theta_{G,i}^t$
                    \EndFor
                    \For{one epoch}
                        \State $\phi_i^{t}\leftarrow\phi_i^{t-1}-\eta\nabla_{\phi_i}\ell_{sup}$
                    \EndFor
		        \State Client $i$ sends $\theta_{G,i}^t$ to the server. 
		    \EndFor
		    \State Server calculates $\theta_G^{t}$ by $\theta_G^{t} \leftarrow \sum_{i\in S_t} \frac{k_i}{\sum_{j\in S_t} k_j} \theta_{G,i}^t$.
		\EndFor
		\\
		\Return $w_1,\dots,w_n$
	\end{algorithmic}
\end{algorithm}
%%%%%%%%%%%%%%%%%%%%%%%%%%%%%%%%%%%%%%%%%%%%%%%%%%%%%%%%%%%%%

\subsection{Discussion}

In this subsection, we discuss FedAFK from three perspectives: communication efficiency and privacy preserving, computation overhead, and model inference.

\textbf{Communication Efficiency and Privacy Preserving.} Our proposed FedAFK transmits only the parameters of the global feature extractor between the server and clients, rather than the entire set of model parameters. This reduction in parameter transmission not only decreases communication overhead but also enhances privacy in federated learning. In deep learning, the feature extractor serves as a foundational component, while the classification head is highly task-specific~\cite{yosinski2014transferable}. By sharing only the feature extractor, we can more effectively safeguard user privacy. Additionally, privacy-preserving techniques such as differential privacy can be integrated into FedAFK to further enhance the system's reliability.

\textbf{Computation Overhead.}  For clients, our proposed FedAFK introduces three additional computation overhead compared to FedAvg~\cite{fedavg}: computing the knowledge transfer loss, updating the personalized model, and calculating the adaptive weight coefficient. In cross-silo scenarios, these costs are tolerable in order to achieve a well-performing personalized model. On the server side, FedAFK only aggregates the updated feature extractors from clients, which helps reduce computation overhead.

\textbf{Model Inference.} In FedAFK, each client obtains a personalized model for inference after training, but we do not provide a global model that can be used for inference. However, achieving this is possible with a few adjustments: by training a global classifier head alongside the global feature extractor, a global inference model can be created. It is worth mentioning that training an additional global classifier head may reduce some of the benefits related to communication efficiency and privacy preservation.

%%%%%%%%%%%%%%%%%%%%%%%%%%%%%%%%%%%%%%%%%%%%%%%%%%%%%%%%%%%%%%%%%%%%%%%%

%%%%%%%%%%%%%%%%%%%%%%%%%%%%%%%%%%%%%%%%%%%%%%%%%%%%%%%%%%
\begin{table*}[h]
  \centering
    \begin{minipage}{0.48\linewidth}
    \centering
    \resizebox{\linewidth}{!}{
    \begin{tabular}{l|*{7}{c}}
    \toprule
    \multirow{2}{*}{FedAFK} & \multicolumn{7}{c}{$\mu=0.5$} \\ 
    \cmidrule{2-8}
    & $\lambda=0.1$ & $\lambda=0.2$ & $\lambda=0.3$ & $\lambda=0.4$ & $\lambda=0.5$ & $\lambda=0.6$ & $\lambda=0.7$ \\ 
    \midrule
    Test Accuracy. & 56.53 & 57.50 & \textbf{57.88} & 57.63 & 57.69 & 57.68 & 57.20\\
    \bottomrule
    \end{tabular}}
    \caption{The test accuracy (\%) for different $\lambda$ values with $\mu=0.5$ for FedAFK on Cifar100 in the default practical heterogeneous setting.}
    \label{tab:eff_left}
  \end{minipage}
  \hfill
  \begin{minipage}{0.48\linewidth}
    \centering
    \resizebox{\linewidth}{!}{
    \begin{tabular}{l|*{7}{c}}
    \toprule
    \multirow{2}{*}{FedAFK} & \multicolumn{7}{c}{$\lambda=0.3$} \\ 
    \cmidrule{2-8}
    & $\mu=0$ & $\mu=0.1$ & $\mu=0.3$ & $\mu=0.5$ & $\mu=0.7$ & $\mu=0.9$ & $\mu=1.0$\\  
    \midrule
    Test Accuracy. & 57.82 & 57.26 & 57.51 & \textbf{57.88} & 57.56 & 46.24 & 45.84\\
    \bottomrule
    \end{tabular}}
    \caption{The test accuracy (\%) for different initial $\mu$ values with $\lambda=0.3$ for FedAFK on Cifar100 in the default practical heterogeneous setting.}
    \label{tab:eff_right}
  \end{minipage}
\end{table*}
%%%%%%%%%%%%%%%%%%%%%%%%%%%%%%%%%%%%%%%%%%%%%%%%%%%%%%%%%%
%%%%%%%%%%%%%%%%%%%%%%%%%%%%%%%%%%%%%%%%%%%%%%%%%%%%%%%%%%%%%%%%%%%%%%%%%%%%%%%%%%%%%%%%
\begin{table*}[h]
    \centering
    \resizebox{0.7\linewidth}{!}{
      \begin{tabular}{p{1.6cm}|p{1.2cm}p{1.2cm}p{1.2cm}|p{1.2cm}p{1.2cm}p{1.2cm}}
      \toprule
      Settings & \multicolumn{3}{c|}{Pathological heterogeneous setting} & \multicolumn{3}{c}{Practical heterogeneous setting} \\
      \midrule
      & MNIST & Cifar10 & Cifar100 & MNIST & Cifar10 & Cifar100\\
      \midrule
      FedAvg & 97.82 & 55.09 & 29.45 & 98.81 & 59.16 & 28.19\\
      FedProx & 97.95 & 55.06 & 30.42 & 98.82 & 59.21 & 28.61\\
      \midrule
      Per-FedAvg & 99.63 & 89.15 & 36.64 & 98.90 & 87.74 & 31.88\\
      pFedMe & 99.75 & 89.41 & 61.23 & 99.42 & 88.09 & 47.67\\
      FedAMP & 99.78 & 89.40 & 61.80 & 99.47 & 88.70 & 46.48\\
      Ditto & 99.80 & 89.77 & 66.85 & 99.54 & 89.41 & 54.02\\
      FedPer & 99.79 & 89.98 & 65.04 & 99.47 & 89.22 & 51.73\\
      FedRep & 99.77 & 89.93 & 61.05 & 99.48 & 90.03 & 51.03\\
      FedROD & 99.85& 90.01& 65.75& 99.56& 89.93& 53.79\\
      FedFomo & 99.83& 89.52& 55.18& 99.33& 88.06& 40.70\\
      APFL & 99.84& 89.56& 61.31& 99.44& 89.41& 50.07\\
      FedPHP & 99.73& 89.97& 65.19& 99.51& 88.92& 54.07\\
      FedALA & 99.86& 89.93& 67.23& 99.56& 89.76& 53.51\\
      \midrule
      FedAFK & {\bf 99.87} & {\bf 90.08} & {\bf 69.18} & {\bf 99.58} & {\bf 90.18} & {\bf 57.88}\\
      \bottomrule
      \end{tabular}}
    \caption{The test accuracy (\%) in the pathological heterogeneous setting and practical heterogeneous setting.}
    \label{tab:ExpResults}
  \end{table*}
%%%%%%%%%%%%%%%%%%%%%%%%%%%%%%%%%%%%%%%%%%%%%%%%%%%%%%%%%%%%%%%%%%%%%%%%%%%%%%%%%%%%%%%%

\section{Experiments}

\subsection{Experimental Setup}

\textbf{Datasets and models.}
To evaluate our proposed FedAFK, we consider image classification tasks and utilize three popular benchmark datasets: MNIST~\cite{mnist}, Cifar-10 and Cifar-100~\cite{cifar}. We consider a multi-layer CNN for both MNIST and Cifar-10, and ResNet18~\cite{resnet} for Cifar-100. The CNN consists of two convolutional layers and two fully connected layers, and the ResNet18 model used is not pre-trained.

\textbf{Data Partitioning.}
Following FedALA~\cite{fedala}, we simulate the Non-IID settings with two typical scenarios, \textit{i.e.}, the pathological heterogeneous setting~\cite{fedavg,shamsian2021personalized} and the practical heterogeneous setting~\cite{lin2020ensemble,moon}. For the pathological setting, we sample data with label amount 2/2/10 for each client on MNIST/Cifar-10/Cifar-100 from a total of 10/10/100 categories, with disjoint data and unbalanced data samples. For the practical setting,  we use the Dirichlet function, denoted as $Dir(\beta)$, to sample data. A smaller $\beta$ indicates a higher degree of heterogeneity in the data and we set $\beta=0.1$ by default. On each client, we divide the data into 75\% and 25\% to form the training dataset and test dataset, respectively.

\textbf{Comparison Baselines.}
To show the superiority of our proposed FedAFK, we compare it with thirteen SOTA FL baselines including FedAvg~\cite{fedavg}, FedProx~\cite{fedprox}, Per-FedAvg~\cite{per-fedavg}, pFedMe~\cite{pfedme}, FedAMP~\cite{fedamp}, Ditto~\cite{ditto}, FedPer~\cite{fedper}, FedRep~\cite{fedrep}, FedROD~\cite{fedrod}, FedFomo~\cite{fedfomo}, APFL~\cite{apfl}, FedPHP~\cite{fedphp}, and FedALA~\cite{fedala}.

\textbf{Implementation Details.}
We implement FedAFK using PyTorch 1.8~\cite{paszke2019pytorch} and conduct experiments on a server equipped with an Intel Xeon Gold 5118 CPU, 96G memory and a NVIDIA RTX A4000 GPU. Following FedAvg~\cite{fedavg}, we set the local batch size to 10 and the local training epochs to $E=1$. We set the number of communication rounds to $T=1000$ to ensure that all methods empirically converge. Following pFedMe~\cite{pfedme}, 
FedFomo~\cite{fedfomo}, and FedALA~\cite{fedala}, we set the number of clients to $n=20$ and the client participation ratio to $\rho=1$. We report the test accuracy of the best global model for traditional FL methods and the average test accuracy of the best local models for personalized FL methods. 

\subsection{Effect of Hyperparameters}
% 分析FedAFK中的两个超参数\mu \lambda的影响.
We vary the values of $\lambda$ to simulate different levels of knowledge transfer. We evaluate the Cifar-100 dataset in the practical heterogeneous setting and report the results in Table~\ref{tab:eff_left}. Choosing a reasonable $\lambda$ can effectively balance the personalization and generalization of the model. FedAFK achieves excellent performance when $\lambda=0.3$.  When $\lambda < 0.3$, the test accuracy decreases due to reduced absorption of global knowledge by the personalized model. Conversely, a larger $\lambda$ leads to excessive alignment of the personalized model with the global model, thereby diminishing its personalization. Hence, we set $\lambda=0.3$ for FedAFK.

Additionally, we evaluate our proposed FedAFK with various initial value of $\mu$. As shown in Table~\ref{tab:eff_right}, when $\mu \leq 0.7$, a relatively high test accuracy is achieved, with the highest accuracy occurring at $\mu = 0.5$. This occurs because, during training, $\mu$ can be adaptively adjusted to obtain an optimal combination of the global feature extractor and local feature extractors. When $\mu > 0.7$, the test accuracy drops significantly, indicating that an excessively large initial value causes severe client-drift\cite{scaffold} in the early stages, making it difficult to correct later. Therefore, we set the initial value of $\mu$ to 0.5 for the subsequent experiments.

\subsection{Performance Comparison}

The main experimental results are presented in Table~\ref{tab:ExpResults}, it is obvious that our proposed method FedAFK obtains the best performances in both pathological heterogeneous setting and practical heterogeneous setting, demonstrating the effectiveness and benefits of adaptive feature aggregation and knowledge transfer. In tasks that are relatively challenging, FedAFK demonstrates significant advantages. In the following, we compare FedAFK with baselines and analyze why FedAFK outperforms all baselines.

The poor performance of traditional FL methods, namely FedAvg and FedProx, can be attributed to their objective of training a global model without incorporating personalized information from the features. In contrast, FedAFK effectively balances both generalization and personalization, providing each client with a personalized model.

Among pFL methods, Per-FedAvg performs poorly because it aggregates a global learning trend for all clients, which does not align with the individual learning trends of each client. In contrast, FedAFK learns a personalized feature extractor for each client through adaptive feature aggregation to align with local learning trends. 

pFedMe, FedAMP, and Ditto all introduce regularization terms to enable local models to learn specific knowledge for enhancing personalization capabilities. Among them, pFedMe learns knowledge from the local model, FedAMP learns from a client-specific server model, and Ditto learns from the global model. Due to the more generic knowledge learned by Ditto, it performs better. Given the effectiveness of this "knowledge transfer" strategy, FedAFK also employs it. Unlike the aforementioned three methods, FedAFK transfers knowledge from feature representations rather than model parameters, which is more direct and effective.

FedPer, FedRep, and FedROD all decouple their model. The former two only share the feature extractor while keeping the classifier head locally, thus preserving global knowledge but being somewhat coarse. FedROD learns two heads with two objectives, however, the two objectives are competing, which adversely affects the pFL task. FedAFK, with a focus on the pFL task, naturally introduces global information through adaptive feature aggregation and knowledge transfer, thus achieving the best performance.

FedFomo aggregates local models using client-specific weights, which results in the loss of some global knowledge. APFL aggregates the entire model through adaptive weights, maintaining both a local model and a mixed model locally, which incurs additional storage costs. Similar to APFL, our proposed FedAFK also has an adaptive weight coefficient $\mu$, but we only maintain one personalized model locally and use $\mu$ to aggregate the feature extractor. Our goal is to obtain an effective feature extractor to assist in building personalized models, rather than directly acquiring personalized models like APFL. FedPHP defines an aggregation parameter that varies with the number of selections and communication rounds, which is insufficient for capturing the connection between the global model and the local models. FedALA provides a more fine-grained model aggregation, but we have found that it is highly sensitive to training hyperparameters. 

%%%%%%%%%%%%%%%%%%%%%%%%%%%%%%%%%%%%%%%%%%%%%%%%%%%%%%%%%%%%%%%%%%%%%%%%%%%%%%
\subsection{Computation Overhead}
We record the mean computation time of all clients in one round on Cifar-100 in the pathological setting, as shown in Table~\ref{tab:comparison}. Our proposed method FedAFK outperforms pFedMe, Ditto and FedRep in terms of computation overhead. However, compared to most pFL methods, \textit{e.g.}, Per-FedAvg, FedAMP, FedPer, FedROD, FedFomo, APFL, FedPHP, and FedALA, FedAFK incurs higher computation costs in each communication round due to the additional training of local models. Therefore, FedAFK is more suitable for clients with relatively sufficient computation power, \textit{e.g.}, enterprises, in cross-silo FL scenarios.
%%%%%%%%%%%%%%%%%%%%%%%%%%%%%%%%%%%%%%%%%%%%%%%%%%%%%%%%%%%%%%%%%%%%%%%%%%%%%%
\subsection{Communication Overhead}
In Table~\ref{tab:comparison}, we also show the communication overhead for one client in one communication round. The communication overhead of the vast majority of FL methods is the same as FedAvg, which uploads and downloads the complete model parameters. FedFomo incurs the highest communication overhead, because it downloads $C$ client models in each round, with $C$ set to 20 in our experiments. Similar to FedPer and FedRep, our proposed method FedAFK only transmits the feature extractor and keeps the classifier head locally, resulting in reduced communication overhead. 

\begin{table}[h]
  \centering
  \resizebox{0.9\linewidth}{!}{
    \begin{tabular}{l|c|c}
    \toprule
    & Computation & Communication\\
    \midrule
    & Time/iter. & Param./iter. \\
    \midrule
    FedAvg & 83s &  $2 * \Sigma$ \\
    FedProx & 89s & $2 * \Sigma$\\
    \midrule
    Per-FedAvg & 146s & $2 * \Sigma$\\
    pFedMe & 307s & $2 * \Sigma$\\
    FedAMP & 90s & $2 * \Sigma$\\
    Ditto & 223s & $2 * \Sigma$\\
    FedPer & 80s & $2 * \alpha_f * \Sigma$\\
    FedRep & 206s & $2 * \alpha_f * \Sigma$\\
    FedROD & 84s & $2 * \Sigma$\\
    FedFomo & 148 & $(1 + C) * \Sigma$\\
    APFL & 133s & $2 * \Sigma$\\
    FedPHP & 163s & $2 * \Sigma$\\
    FedALA & 87s & $2 * \Sigma$\\
    \midrule
    FedAFK & 178s & $2 * \alpha_f * \Sigma$\\
    \bottomrule
    \end{tabular}}
  \caption{Mean computation time (s) of all clients on Cifar-100 and the communication overhead (transmitted parameters per iteration). $\Sigma$ is the parameter amount in the backbone. $\alpha_f$ ($\alpha_f < 1$) denotes the ratio of the parameters of the feature extractor in the backbone. $C$ ($C \ge 1$) is the number of the received client models on each client in FedFomo. }
    \label{tab:comparison}
\end{table}

%%%%%%%%%%%%%%%%%%%%%%%%%%%%%%%%%%%%%%%%%%%%%%%%%%%%%%%%%%%%%%%%%%%%%%%%%%%%%%

\subsection{Ablation Studies}
We have two key design components in FedAFK, \textit{i.e.}, adaptive feature aggregation(AFA) and knowledge transfer(KT). To show the effectiveness, we conduct ablation experiments with the following four cases("\textit{w/o}" is short for "without"): (a) \textit{w/o} AFA, (b) \textit{w/o} KT, (c) \textit{w/o} both AFA and KT: since we use local models for inference, FedAFK turns into Standalone,\textit{i.e.}, each client trains its model solely, (d) the complete FedAFK approach. We report the test accuracy of the four ablation cases in Table~\ref{tab:ablation}.
%%%%%%%%%%%%%%%%%%%%%%%%%%%%%%%%%%%%%%%%%%
\begin{table}[h]
    \centering
    \resizebox{\linewidth}{!}{
      \begin{tabular}{l|cccc}
      \toprule
       & \textit{w/o} AFA & \textit{w/o} KT & \textit{w/o} AFA \& KT & FedAFK\\
      \midrule
      MNIST & 99.21 & 99.37 & 99.17 & \textbf{99.58} \\
      Cifar10 & 88.69 & 89.10 & 87.83 & \textbf{90.18}\\
      Cifar100 & 48.61 & 48.36 & 46.17 & \textbf{57.88}\\
      \bottomrule
      \end{tabular}}
    \caption{The test accuracy (\%) of four ablation cases in the practical heterogeneous setting. }
    \label{tab:ablation}
  \end{table}
%%%%%%%%%%%%%%%%%%%%%%%%%%%%%%%%%%%%%%%%%

As demonstrated in Table~\ref{tab:ablation}, both adaptive feature aggregation and knowledge transfer can help improve the test accuracy and the complete FedAFK approach achieves the highest test accuracy. This indicates that our proposed method effectively trains a better personalized feature extractor while balancing generalization and personalization. It is worth noting that, in tasks that are relatively challenging(such as training resnet on Cifar100), the absence of any single design results in a noticeable decrease in test accuracy compared to the complete  FedAFK, which indicates that adaptive feature aggregation and knowledge transfer can mutually facilitate each other.

%%%%%%%%%%%%%%%%%%%%%%%%%%%%%%%%%%%%%%%%%%%%%%%%%%%%%%%%%%%%%%%%%%%%%%%%

\section{Conclusion And Future Work}
In this paper, we propose a novel personalized federated learning method FedAFK, which addresses challenging FL scenarios with statistical heterogeneity. To enhance the representational capability of the feature extractor, FedAFK transfers the global knowledge to the local feature extractor and adaptively aggregates the global feature extractor and the local feature extractor as the personalized feature extractor. Through extensive experiments, we demonstrate that FedAFK outperforms thirteen SOTA methods. As future work, we plan to further improve FedAFK in more complex settings, $\textit{e.g.}$, clients with dynamic data distribution.

%%%%%%%%%%%%%%%%%%%%%%%%%%%%%%%%%%%%%%%%%%%%%%%%%%%%%%%%%%%%%%%%%%%%%%%%

%%% The acknowledgments section is defined using the "acks" environment
%%% (rather than an unnumbered section). The use of this environment 
%%% ensures the proper identification of the section in the article 
%%% metadata as well as the consistent spelling of the heading.

\begin{acks}
If you wish to include any acknowledgments in your paper (e.g., to 
people or funding agencies), please do so using the `\texttt{acks}' 
environment. Note that the text of your acknowledgments will be omitted
if you compile your document with the `\texttt{anonymous}' option.
\end{acks}

%%%%%%%%%%%%%%%%%%%%%%%%%%%%%%%%%%%%%%%%%%%%%%%%%%%%%%%%%%%%%%%%%%%%%%%%

%%% The next two lines define, first, the bibliography style to be 
%%% applied, and, second, the bibliography file to be used.

\bibliographystyle{ACM-Reference-Format} 
\bibliography{sample}

%%%%%%%%%%%%%%%%%%%%%%%%%%%%%%%%%%%%%%%%%%%%%%%%%%%%%%%%%%%%%%%%%%%%%%%%

\end{document}